\documentclass[lettersize,journal]{IEEEtran}
\usepackage{amsmath,amsfonts}
\usepackage{algorithmic}
\usepackage{algorithm}
\usepackage{array}
\usepackage[caption=false,font=footnotesize,labelfont=rm,textfont=rm]{subfig}
\usepackage{textcomp}
\usepackage{stfloats}
\usepackage{url}
\usepackage{verbatim}
\usepackage{graphicx}
\usepackage{cite}
\usepackage{bbding}
\usepackage{booktabs}
\usepackage{multirow}
\usepackage{color}
\usepackage{hyperref}
\usepackage{cleveref}

\def\etal{\textit{et al. }}

\begin{document}

\title{LLPR: Location-aware learning and physics-based reconstruction for raindrop removal from a single image}

\author{Zewei He$^\dagger$$^\ast$, Xingyu Liu$^\dagger$, Xing Luo, Guizhong Fu, Zixuan Chen, Yu Chen, Jinlei Li, \\Zhe-Ming Lu$^\ddagger$,~\IEEEmembership{senior member,~IEEE}
	\thanks{This work was supported in part by the National Natural Science Foundation of China under Grant No. 52305590, in part by the Zhejiang Provincial Natural Science Foundation of China under Grant No. LQ24F010004.}
	\thanks{Z. He is with Huanjiang Laboratory, Zhuji, P.R. China and School of Aeronautics and Astronautics, Zhejiang University, Hangzhou, P.R.China (e-mail: zeweihe@zju.edu.cn).}
	\thanks{X. Liu, X. Luo, Z. Chen, Y. Chen, J. Li and Z.M. Lu are with School of Aeronautics and Astronautics, Zhejiang University, Hangzhou, P.R.China (zheminglu@zju.edu.cn).}
	\thanks{G. Fu is with School of Mechanical Engineering, Suzhou University of Science and Technology, Suzhou, P.R.China (fuguizhongchina@163.com).}
	\thanks{$^\dagger$The first two authors contribute equally to this work.}
	\thanks{$^\ddagger$Zhe-Ming Lu is the team lead.}
	\thanks{$^\ast$Corresponding author: Zewei He (e-mail: zeweihe@zju.edu.cn).}
}

\maketitle

\begin{abstract}
Raindrops can cause occlusion and distortion in the background scenes due to their adherence to windows or camera lenses. Existing raindrop removal methods concentrate on designing sophisticated CNN or Transformer architectures to recover distorted and missing texture. In this paper, we try to integrate location information and physical model into off-the-shelf CNN or Transformer architectures to help improve their performance. Specifically, we notice that existing methods deploy a preprocessing sub-network to generate a binary or soft mask to indicate the raindrop location, which will increase the network parameters and computational complexity. In contrast, a location-aware learning branch is embedded to teach the encoder in the training phase with the capability of perceiving the position of the raindrops. Note that this location-aware learning branch can be removed during the inference process (achieving performance improvements at no cost). Furthermore, instead of directly reconstructing the raindrop-free image (i.e., background scene), we devise a physics-based reconstruction scheme to first learn the transparency matrix and the raindrop layer. The latent background layer is then reversely derived based on the physical model. By combining the above-mentioned components, we propose our location-aware learning and physics-based reconstruction (LLPR) framework for this challenging ill-posed problem. We also collect a real-world raindrop-degraded image dataset, which is challenging for single-image raindrop removal (SIRR) methods. Extensive experimental results demonstrate the effectiveness and generality of our LLPR framework, achieving superior performance against state-of-the-art SIRR methods. The code will be made available upon acceptance. 
\end{abstract}


\section{Introduction}

\begin{figure}[t]
	\centering
	\includegraphics[width=0.99\linewidth]{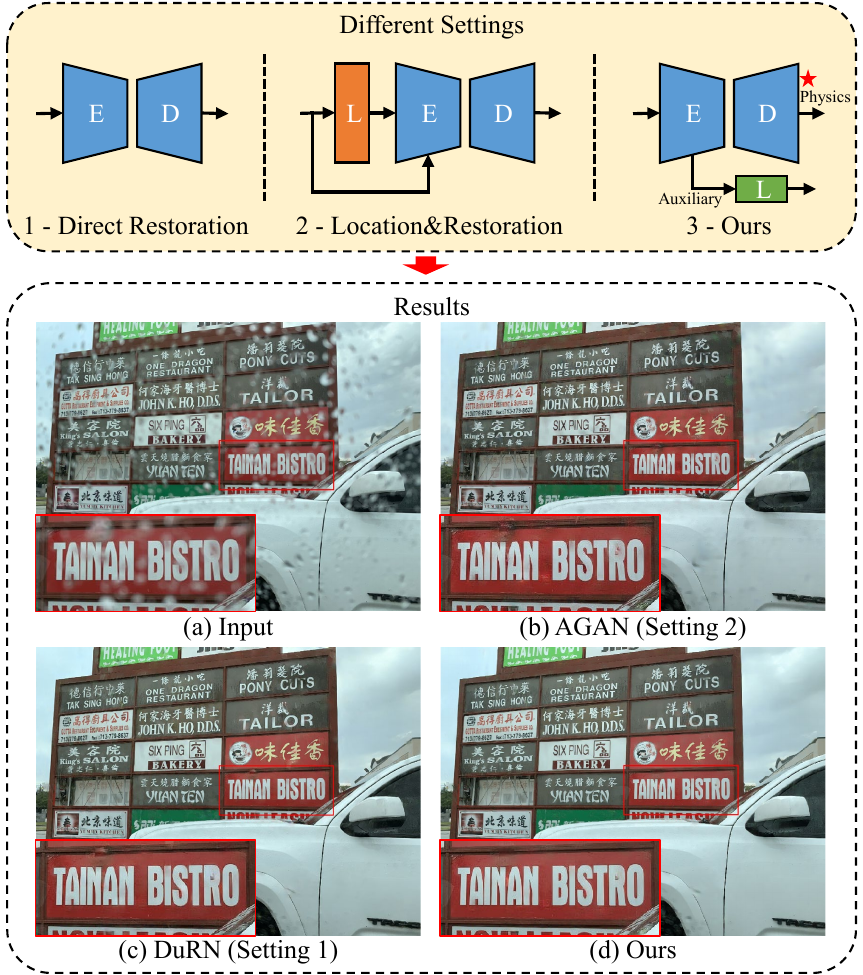}
	\caption{Top: Schematic diagram of different settings. E, D, and L denote encoder, decoder, and location, respectively. Bottom: Comparison of raindrop removal performance. This is a photograph of a billboard taken through an automobile's windshield by an iPhone in a real-world rainy day (wild scene).}
	\label{fig:fig1}
\end{figure}

\IEEEPARstart{I}{n} rainy weather, glass windows, windscreens and lenses are often covered by raindrops with different sizes and shapes.
When capturing background photographs through them, the visibility and quality may be hampered or degraded \cite{Eigen2013,Quan2019ICCV}, producing a significant impact on downstream high-level vision tasks, e.g., object detection, recognition.
With the widespread use of smartphones and outside-mounted surveillance cameras, such situations have become increasingly common \cite{Shao2021Neurocom,Guo2020TIP,Luo2021TCSVT}.
Therefore, developing effective raindrop removal methods is certainly an urgent need for recovering the scenes corrupted by adherent raindrops.

The physical model of an image captured through the window with raindrops is quite complicated \cite{You2016}.
In our paper, we consider a simple linear model by following \cite{You2016,Quan2019ICCV} and the formula can be expressed as:

\begin{equation}
	\textbf{I} = (1 - \textbf{A})\odot \textbf{B} + \textbf{A}\odot \textbf{R},
	\label{Eqn1}
\end{equation}
where $\odot$ denotes the element-wise multiplication, $\textbf{I} \in \mathbb{R}^{3\times H\times W}$ is the captured image with raindrops, $\textbf{B} \in \mathbb{R}^{3\times H\times W}$ refers to background scene (i.e., latent raindrop-free image), $\textbf{R} \in \mathbb{R}^{3\times H\times W}$ represents the raindrop layer, and $\textbf{A} \in \mathbb{R}^{3\times H\times W}$ indicates the transparency matrix.
Each entry of $\textbf{A}$ denotes the proportion of the light path covered by a raindrop.

\begin{figure}[t]
	\centering
	\includegraphics[width=0.99\linewidth]{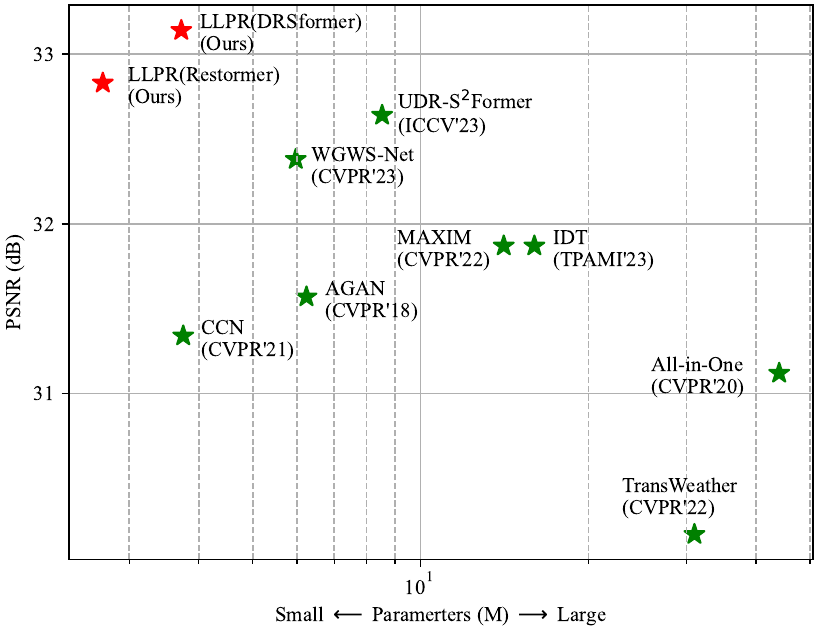}
	\caption{PSNR vs. number of parameters for various algorithms. We compare our LLPR framework with some state-of-the-art raindrop removal methods. The results are measured on Test-a dataset.}
	\label{fig:fig2}
\end{figure}

With the rapid development of deep learning, models of CNN and Transformer have achieved superior performance \cite{Eigen2013,Qian2018CVPR-AGAN,Quan2019ICCV,Peng2020PRL,Shao2021TIP,Chen2023ICCV-UDR}.
There are primarily two configurations for existing methods, as illustrated in Fig.~\ref{fig:fig1}.
Based on the first setting (i.e., direct restoration), Eigen \etal \cite{Eigen2013} pioneerly proposed a three-layer CNN architecture to tackle the single-image raindrop removal (SIRR) problem via directly learning the degraded-to-clear mapping function.
Afterward, some raindrop removal methods have been proposed to enhance performance by designing more sophisticated network architectures \cite{Liu2019CVPR-DuRN,Tu2022CVPR-MAXIM,Sun2024ECCV-Histogram}.
These methods focus on network design and neglect the physical model, making them prone to failure when dealing with complex and varied real-world images.
Another setting contains two steps, i.e., identify the regions of raindrops and then recover the pixels of identified regions.
For example, Qian \etal \cite{Qian2018CVPR-AGAN} proposed the attentive generative adversarial network (AGAN), which learns a raindrop location mask by the Res-LSTM sub-network \cite{He2016CVPR-Resnet,Shi2015NIPS-LSTM}.
Similar strategy was adopted by Shao \etal \cite{Shao2021TIP} with different mask labels.
Quan \etal \cite{Quan2019ICCV} proposed a shape-driven attention (based on edges) which exploits physical shape priors of raindrops (i.e., convexness and contour closedness) to accurately locate the raindrops.
Among these methods, the obtained location mask is regarded as the prior information and integrated into the restoration network.
However, the extraction and integration of prior knowledge inevitably cost extra computational resources.

In this paper, we try to address SIRR problem by shifting our focus towards converting existing prior extractors into cost-free solutions and exploring physics-based techniques without re-designing novel structures.
To start, an auxiliary location-aware learning branch is deployed in the encoder to teach the feature extractor with the capability of perceiving position of the raindrops (refer to the top right of Fig.~\ref{fig:fig1}).
This branch is only effective during the training phase, and can be removed seamlessly during the inference.
Then, we modify the reconstruction part to estimate other key components in Eqn.~\ref{Eqn1} (e.g., transparency matrix $\textbf{A}$ and raindrop effect $\textbf{R}$), and employ them to derive the raindrop-free prediction ($\textbf{A}$ and $\textbf{R}$ are learned without supervision, and only the derived raindrop-free prediction is supervised).
We denote this sophisticated design as physics-based reconstruction scheme.
This scheme can be regarded as a soft constraint on the correlation between $\textbf{A}$ and $\textbf{R}$, which is beneficial for improving the raindrop removal performance in wild scenarios.
Last but not least, with estimated $\textbf{A}$ and $\textbf{R}$, we propose a cycle loss function to pose supervision on the input raindrop-degraded image.
This cycle loss function can be regarded as a self-supervised learning strategy, which is also effective in improving the performance of our LLPR framework.

Note that, above-mentioned techniques can be equipped with different restoration network architectures, and by combining them together, we propose our location-aware learning and physics-based reconstruction (LLPR) framework.
Figure~\ref{fig:fig2} shows the effectiveness and efficiency of our LLPR framework.
The diagram of our LLPR framework is illustrated in Fig.~\ref{fig:fig3}.
Figure~\ref{fig:fig1} shows the superiority of our LLPR framework against state-of-the-art raindrop removal methods in real-world rainy weather (wild scene). 
Our method, i.e., LLPR(DRSformer) experimentally shows strong generalization to wild images.
To conclude, the main contributions of this work are summarized as follows:
\begin{itemize}
	\item An auxiliary location-aware learning branch is devised to endow the encoder with the capability of perceiving the position of the raindrops. A crucial difference from previous location-aware methods \cite{Qian2018CVPR-AGAN,Shao2021TIP} lies in the fact that our auxiliary branch is independent of the backbone network and only take effect during the training phase, without consuming any resources during the testing phase.
	\item We propose a physics-based reconstruction scheme by explicitly considering the physical model of the raindrop formation process. Learning key components like transparency matrix $\textbf{A}$ and raindrop effect $\textbf{R}$ can constrain the correlation of them softly, improving the raindrop removal performance in complex real-world scenarios. Furthermore, we develop a cycle loss with estimated $\textbf{A}$ and $\textbf{R}$ to pose supervision on input image. To the best of our knowledge, this is the first time these techniques are adopted in raindrop removal domain.
	\item The proposed LLPR framework can be flexibly combined with various image restoration networks (e.g., DRSformer \cite{Chen2023CVPR-DRSformer} and Restormer \cite{Zamir2022CVPR-Restormer}) to achieve state-of-the-art performance for SIRR without designing sophisticated CNN or Transformer module. In addition, we collect a wild testing dataset, called Test-wild, which contains 226 challenging raindrop-degraded images captured in real-world rainy weather.
\end{itemize}

\section{Related work}
\label{sec: related work}
We categorize existing SIRR approaches into multi-task methods, restoration methods, and task-specific methods.

\subsection{Multi-task methods}
This category aims to handle multiple bad weather degradations (e.g., rain, raindrop, haze) with a single trained model. 
All-in-One Network \cite{Li2020CVPR} is a very representative approach, which employs multiple task-specific encoders and a shared decoder to realize it.
Valanarasu \etal \cite{Valanarasu2022CVPR} proposed a transformer-based end-to-end model, and a single encoder-decoder architecture is designed to restore images degraded by various weather conditions. 
The model incorporates a transformer encoder that employs intra-patch transformer blocks to enhance attention within individual patches.
Later, Zhu \etal \cite{Zhu2023CVPR} designed a unified framework with a two-stage training strategy to explore the weather-general and weather-specific features automatically.

\subsection{Restoration methods}
This category aims to develop a versatile network architecture capable of handling various forms of image degradation.
They primarily train a series of degradation-aware models with the identical architecture for different types of degradation.
Zamir \etal \cite{Zamir2022CVPR-Restormer} proposed a transformer-based model that utilizes self-attention mechanism to capture long-range dependencies and complex patterns in images. 
Liu \etal \cite{Liu2019CVPR-DuRN} proposed a dual residual network designed to handle a wide range of image degradation tasks. 
Tu \etal \cite{Tu2022CVPR-MAXIM} investigated a model that leverages a multi-axis MLP architecture to generalize across multiple image restoration tasks.
Huang \etal \cite{Huang2024TMM-WaveDM} proposed a Wavelet-Based Diffusion Model (WaveDM) to learn the distribution of clean images in the wavelet domain conditioned on the wavelet spectrum of degraded images after wavelet transform.

\subsection{Task-specific methods}
This category focuses on rain-related degradation, which is more specific than the above two categories.
Eigen \etal \cite{Eigen2013} firstly introduced deep learning for SIRR problem. The main idea of the method involves training a shallow CNN and using pairs of raindrop-degraded and corresponding raindrop-free images. 
Qian \etal \cite{Qian2018CVPR-AGAN} applied an attentive generative network using adversarial training, combining the generative adversarial network (GAN) and attention mechanism to recover Raindrop-free image from degraded images.
Quan \etal \cite{Quan2019ICCV} proposed a convolutional neural network with double attention mechanism to enhance the removal process. This mechanism combines shape-driven attention, which utilizes physical shape priors such as convexity and contour closedness to precisely locate raindrops, with channel re-calibration to improve robustness across varying raindrop appearances.
Shao \etal \cite{Shao2021TIP} introduced a multi-scale attention network, which incorporates a soft mask that encodes the blurring level of raindrops, guiding the network to focus on varying degrees of blur. Additionally, it utilizes a multi-scale fusion representation through a pyramid structure, coupled with an iterative mechanism to address raindrops at different scales while reducing the impact of redundant noise.

More recently, there has been an emergence of networks that are designed to address both raindrop degradation and rain streak degradation simultaneously \cite{Quan2021CVPR-CCN,Xiao2023PAMI,Chen2023ICCV-UDR}.
Quan \etal \cite{Quan2021CVPR-CCN} proposed a complementary cascaded network (CCN) architecture to remove rain streaks and raindrops in a unified framework.
Xiao \etal \cite{Xiao2023PAMI} presented a transformer system comprising a complementary window-based transformer and spatial transformer, enabling improved capture of short-range and long-range dependencies in rainy scenes.
Chen \etal \cite{Chen2023ICCV-UDR} designed a sparse sampling transformer based on uncertainty-driven ranking strategy. The model samples relevant image degradation information, enabling the effective modeling of underlying degradation  

\section{Methodology}
\label{sec: methodology}

\begin{figure*}[t]
	\centering
	\includegraphics[width=0.99\linewidth]{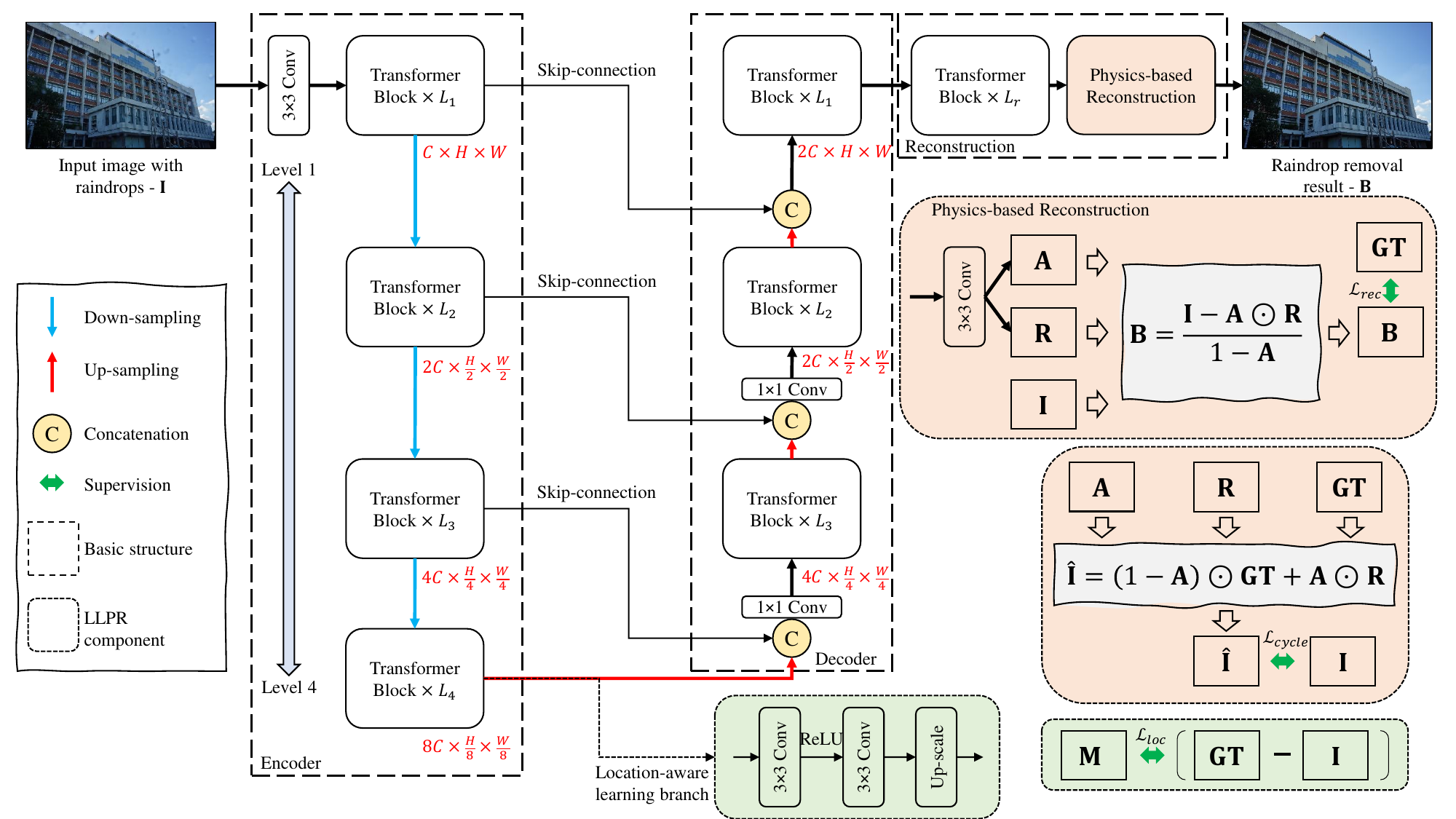}
	\caption{The diagram of our location-aware learning and physics-based reconstruction (LLPR) framework. It mainly consists of a four-level U-net-like backbone. The rectangular boxes with background colors represent the core modules.}
	\label{fig:fig3}
\end{figure*}

\subsection{Basic architecture}
As shown in Fig.~\ref{fig:fig3}, our location-aware learning and physics-based reconstruction (LLPR) framework mainly consists of a U-net-like backbone.
The backbone is a four-level encoder-decoder-like structure, which contains three parts: encoder, decoder, and reconstruction.
There are three down-sampling operations and three corresponding up-sampling operations.
The down-sampling operation halves the spatial dimensions and doubles the number of channels.
It is realized through a convolution layer and a pixel-unshuffle layer \cite{Chen2019CVPRW}.
The up-sampling operation can be regarded as the inverse form of the down-sampling operation, which is realized through a convolution layer and a pixel-shuffle operation \cite{Shi2016CVPR}.
The dimensional sizes from Level 1 to 4 are $C\times H\times W$, $2C\times \frac{H}{2}\times \frac{W}{2}$, $4C\times \frac{H}{4}\times \frac{W}{4}$, and $8C\times \frac{H}{8}\times \frac{W}{8}$, respectively.
Given a raindrop-degraded image $\textbf{I}$, the ultimate goal is to predict the raindrop-removal result $\textbf{B}$.
Note that we do not design sophisticated Transformer blocks in Fig.~\ref{fig:fig3}, and instead, choose the off-the-shelf blocks.

\subsection{Location-aware learning branch}
In practice, the spatial regions occluded by raindrops are not explicitly provided, which hold significant importance for addressing single-image raindrop removal (SIRR) problem.
The regions (i.e., location information) can offer essential clues for effective restoration of the background scene and unambiguous removal of degraded/corrupted effect brought by the raindrops.

Some researches have noticed this point and started to learn the location information of raindrops within the network \cite{Qian2018CVPR-AGAN,Shao2021TIP,Quan2019ICCV,Liu2024Neurocom}.
For example, Qian \etal \cite{Qian2018CVPR-AGAN} designed a binary mask (via a threshold) to determine whether a pixel is part of a raindrop region.
This binary mask is then used to supervise the learning of the 2D attention map.
Later, Shao \etal \cite{Shao2021TIP} proposed a soft mask with values ranging from -1 to 1 to indicate the blurring level of the raindrops on the background.
Both methods build a sub-network (containing Res-LSTM blocks \cite{He2016CVPR-Resnet,Shi2015NIPS-LSTM}) and introduce an iterative mechanism \textbf{before} the restoration network to learn a spatial attentive map $\textbf{S} \in \mathbb{R}^{H\times W}$.

However, we argue that separating the tasks of location extraction and restoration is not an optimal approach.
On one hand, how to employ the learned spatial attention map to guide the restoration network remains largely under-explored.
Fundamental integration methods including addition, multiplication, and concatenation may not guarantee that location information can be fully utilized by the restoration network.
On the other hand, they will inevitably increase additional computation burden in the training or testing phase.

Therefore, we build an auxiliary location-aware learning branch which can teach the feature extractors (i.e., Transformers block in Fig.~\ref{fig:fig3}) with the capability of perceiving the position of the raindrops, integrating it into the encoder during the training phase.  
This scheme enhances the encoder's understanding of contextual information, thereby facilitating more accurate raindrop-relevant feature learning.

Specifically, a location-aware learning branch with two $3\times 3$ convolution layers and a ReLU activation layer is connected with the latent features in level 4 to learn an initial location map $\textbf{M}^* \in \mathbb{R}^{3\times \frac{H}{8}\times \frac{W}{8}}$.
The first convolution layer compresses the number of channels to half of the original, while the second one outputs three-channel $\textbf{M}^*$.
To facilitate the calculation of gradients, $\textbf{M}^*$ is up-sampled to $\textbf{M} \in \mathbb{R}^{3\times H\times W}$ through the bilinear interpolation method.
Note that, during the training phase, the difference between $\textbf{GT}$ and $\textbf{I}$ is utilized as the label to supervise the learned location map $\textbf{M}$.
Certainly, one could also employ raindrop location masks to supervise this branch.
However, acquiring precise raindrop location masks is extremely time-consuming and requires substantial human efforts.
This simple strategy is sufficient for our purpose.
Following \cite{Qian2018CVPR-AGAN,Quan2019ICCV,Shao2021TIP}, mean squared error (MSE) is adopted to drive the learning of this auxiliary branch.

\begin{equation}
	\mathcal{L}_{loc} =\dfrac{1}{N}\sum^{N} ||\textbf{M} - (\textbf{GT} - \textbf{I})||_2^2,
\end{equation}
where $N$ denotes the number of elements.
We omit the subscripts in order to simplify the notation of the formula.
Once the training is completed, this branch can be seamlessly removed.
This also implies that it will not incur extra inference time during the testing phase.

\begin{figure*}[h]
	\centering
	\includegraphics[width=0.99\linewidth]{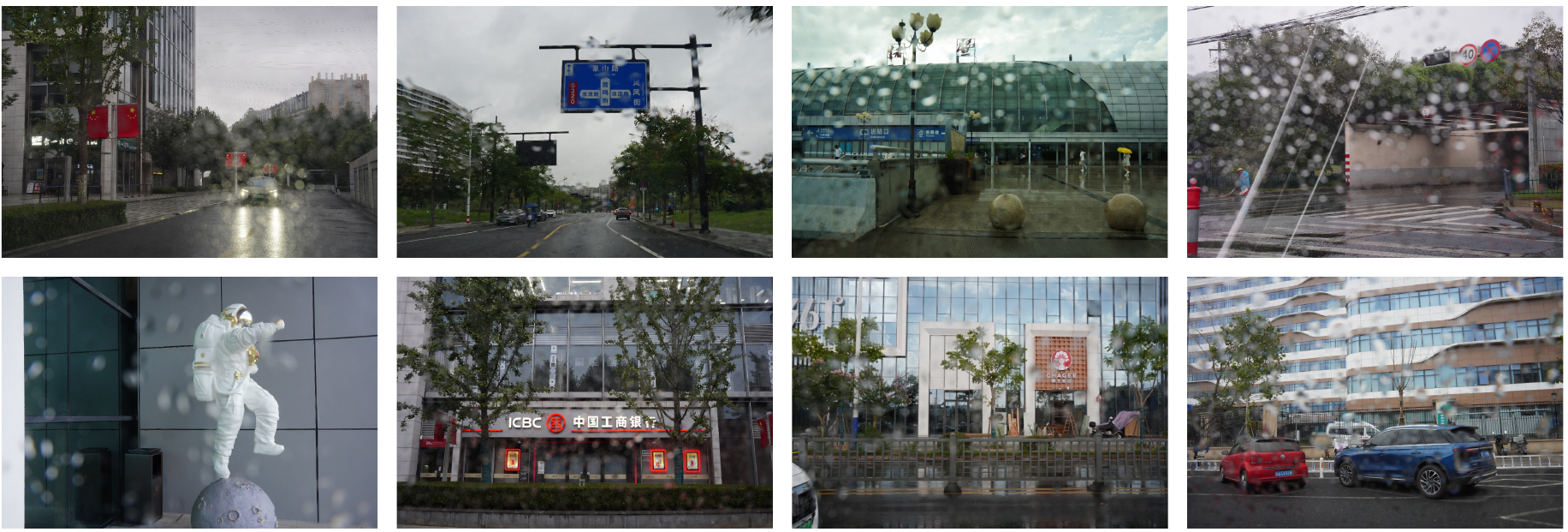}
	\caption{Image samples of Test-wild dataset which are captured under actual rainy weather conditions with diverse scenarios.}
	\label{fig:fig4}
\end{figure*}

\subsection{Physics-based reconstruction scheme}
According to Eqn.~\ref{Eqn1}, directly estimating the latent raindrop-free image $\textbf{B}$ from the observed raindrop-degraded image $\textbf{I}$ is a very challenging task, since \textbf{A} and \textbf{R} are not pre-defined (this problem is highly ill-posed).
However, almost all existing methods hold the same assumption that the raindrop removal operation can be learned through a vast amount of raindrop-degraded and corresponding raindrop-free training image pairs (i.e., $\{\textbf{I}_i, \textbf{GT}_i\}$) \cite{Eigen2013,Peng2020PRL}.
The performance of the learned CNN or Transformer models for raindrop removal is guaranteed by the architecture designs or extracted information, like edge, uncertainty, blurring level \cite{Quan2019ICCV,Shao2021TIP,Chen2023ICCV-UDR}.
They are not universally applicable and may fail under specific conditions.

As illustrated in the right side of Fig.~\ref{fig:fig3}, another unexplored technical path is to estimate other key components in Eqn.~\ref{Eqn1} (e.g., $\textbf{A}$ and $\textbf{R}$) and then derive the $\textbf{B}$ through following formula:

\begin{equation}
	\textbf{B} = \dfrac{\textbf{I} - \textbf{A}\odot \textbf{R}}{1-\textbf{A}}.
	\label{Eqn.3}
\end{equation}

Accordingly, we use a $3\times 3$ convolution layer to generate a feature bank ($\mathbb{R}^{6\times H\times W}$), and then split it into $\textbf{A}$ ($\mathbb{R}^{3\times H\times W}$) and $\textbf{R}$ ($\mathbb{R}^{3\times H\times W}$).
Differs from previous physics-based low-level methods \cite{Li2019CVPR,Zhang2018CVPR} that impose strict supervision on the estimated components, we only supervise the derived $\textbf{B}$. A potential reason may lie in the absence of labels for \textbf{A} and \textbf{R}.
As a result, the network architecture introduces physical priors and \textbf{softly} constrains the correlation between $\textbf{A}$ and $\textbf{R}$.
We refer to Eqn.~\ref{Eqn.3} as the physics-based reconstruction scheme.

By explicitly considering the physical model of the raindrop formation process, our physics-based reconstruction scheme can better remove raindrops under actual conditions (please refer to Fig.~\ref{fig:fig1}).
We adopt mean absolute error (MAE) to update parameters.
\begin{equation}
	\mathcal{L}_{rec} = \dfrac{1}{N}\sum^{N} ||\textbf{B} - \textbf{GT} ||_1,
\end{equation}

\subsection{Cycle loss function}
Though using raindrop-free image $\textbf{GT}$ to directly or indirectly guide the prediction $\textbf{B}$ is very straightforward and effective, few researchers pay attention to input $\textbf{I}$.
We surprisingly find that given estimated $\textbf{A}$ and $\textbf{R}$, realistic raindrop-degraded image $\hat{\textbf{I}}$ can be synthesized as:

\begin{equation}
	\hat{\textbf{I}} = (1 - \textbf{A})\odot \textbf{GT} + \textbf{A}\odot \textbf{R}.
	\label{Eqn5}
\end{equation}

If the estimates of $\textbf{A}$ and $\textbf{R}$ are sufficiently accurate, the simulated image $\hat{\textbf{I}}$ according to Eqn.~\ref{Eqn5} can infinitely approximate $\textbf{I}$.
Based on this, we propose a cycle loss function $\mathcal{L}_{cycle}$ to pose supervision on $\hat{\textbf{I}}$, thereby facilitating the learning of $\textbf{A}$ and $\textbf{R}$.

\begin{equation}
	\mathcal{L}_{cycle} = \dfrac{1}{N}\sum^{N} ||\hat{\textbf{I}} - \textbf{I} ||_1.
\end{equation}

Finally, the total loss function $\mathcal{L}$ can be defined as:
\begin{equation}
	\mathcal{L} = \mathcal{L}_{rec} + \lambda_1\mathcal{L}_{loc} + \lambda_2\mathcal{L}_{cycle},
\end{equation}
where $\lambda_1$ and $\lambda_2$ denote the weights for balancing these losses. In our implementation, $\lambda_1$ and $\lambda_2$ are set to 1 and 0.1, respectively.

\subsection{Test-wild: raindrop-degraded images in rainy weather}
Surprisingly, we find that there is rarely publicly available dataset for testing SIRR methods in wild scenarios, i.e., genuine raindrops in rainy weather.
In previous datasets \cite{Qian2018CVPR-AGAN,Quan2021CVPR-CCN}, the raindrop-degraded images are either synthesized by adding raindrop effects to the background images or artificially created by spraying water onto the glass surfaces.
As shown in Fig.~\ref{fig:fig4}, we collect some challenging real-world raindrop-degraded images in rainy weather via mobile phones or commercial cameras to constitute the Test-wild dataset.
Note that Test-wild is intended solely for testing purposes, as it does not include corresponding aligned ground truths.
Although the Qian's dataset \cite{Qian2018CVPR-AGAN} is also collected in reality (not synthetic), their shooting setting is relatively ideal and quite different from the glass window or automobile's windshield in real situations.
For example, the distance between the camera and the glass remains relatively constant (2 to 5 cm), and the glass they used is flat without curvature.
Moreover, their raindrops are created by spraying water onto the glass, remaining subtle differences compared with raindrops observed in the wild.

We hold the camera behind the windshield of a vehicle or behind a glass window to capture Test-wild data in rainy days.
The raindrops captured in these images are not artificially created, but rather naturally formed in the wild.
To enhance the diversity of raindrops, we randomly employ different camera poses by adjusting the angle and distance.
Similar to \cite{Qian2018CVPR-AGAN}, we control the focus of the camera to make it on the background scene.
In total, 226 images are captured with various background scenes \footnote{Note that this dataset will continue to be expanded.}.
Sony ILCE-7RM4A, Nikon D7100, and several iPhones are used for the image acquisition.
We hope these data can contribute to the advancement and development of the SIRR task, and benefit the entire community.

\section{Experiments}
\label{sec: experiment}

\subsection{Experimental configuration}
\subsubsection{Datasets} 
We train and test our LLPR framework on Qian's dataset \cite{Qian2018CVPR-AGAN} which contains over 1100 image pairs, and each pair is made up of a raindrop-degraded image and corresponding raindrop-free image.
Following \cite{Qian2018CVPR-AGAN,Quan2019ICCV,Shao2021TIP}, we use 861 image pairs in the training phase, and the rest for testing. 
There are two datasets, i.e., Test-a and Test-b, which include 58 and 249 image pairs respectively \footnote{The image pairs in Test-a are better-aligned than those in Test-b, thus we perform ablation study on Test-a instead of Test-b.}.
In addition, we also test on our collected Test-wild dataset.

\subsubsection{Evaluation metrics} 
Peak Signal-to-Noise-Ratio (PSNR) and Structural Similarity Index Measurement (SSIM) \cite{Wang2004TIP-SSIM}, which are commonly used to measure the image quality among the computer vision community, are employed for SIRR performance evaluation.
For fair comparisons with other methods \cite{Chen2023ICCV-UDR,Qian2018CVPR-AGAN,Shao2021TIP}, the metrics are calculated on the luminance channel of YCbCr color space.

\subsubsection{Implementation details} 
Our LLPR framework is implemented on PyTorch deep learning platform \cite{Paszke2019NIPS-PyTorch} with a single NVIDIA RTX4090 GPU.
For Transformer blocks, we select the basic module used in DRSformer \cite{Chen2023CVPR-DRSformer} and Restormer \cite{Zamir2022CVPR-Restormer}.
The number of blocks deployed in each level $L_{1:4}$ is set to $\{2,4,4,8\}$ and the channel number $C$ is set to 16.
For DRSformer, the refinement block number $L_r$ is 0, while for Restormer, it is set to the same value with $L_1$.
The number of heads in self-attention at different levels are set to $\{1,2,4,8\}$.
The channel expansion ratio in feed-forward network is designated to 2.66.
The entire network is optimized with the AdamW \cite{AdamW} optimizer and $\beta_1$, $\beta_2$, $\varepsilon$ are set to default values, i.e., 0.9, 0.999, $1e^{-8}$.
Moreover, we train models with a batch size of 4 and with an initial learning rate of $3e^{-4}$ for the first 100K iterations, which will gradually reduced to $1e^{-6}$ using cosine annealing schedule \cite{He2019CVPR-Bag} during the remaining 200K iterations (there is a total of 300K iterations).
During the training phase, patches of size $256\times 256$ are randomly cropped from the training dataset, and we apply horizontal and vertical flips as the techniques for data augmentation.

\subsection{Ablation study}
\subsubsection{Overall}
To validate the effectiveness of our proposed LLPR, we construct four models:
(1) Using the basic architecture described in Sec.~\ref{sec: methodology}. We denote it as Baseline.
(2) Adding physics-based reconstruction scheme to Baseline. We denote it as (1)+PR.
(3) Upon the basis of (2), further adding location-aware learning branch. We denote it as (2)+LL. It can also be equivalently represented as (1)+PR+LL.
(4) Upon the basis of (3), further adding the cycle loss function $\mathcal{L}_{cycle}$. We denote it as (3)+$\mathcal{L}_{cycle}$. This is the whole LLPR framework in this paper.
We adopt two different basic architectures, within which the Transformer blocks originate from DRSformer \cite{Chen2023CVPR-DRSformer} and Restormer \cite{Zamir2022CVPR-Restormer}.

\begin{table}[ht]
	\footnotesize
	\centering
	\caption{Ablation study of the key components (physics-based reconstruction scheme, location-aware learning branch, and cycle loss function $\mathcal{L}_{cycle}$) of our LLPR framework.}
		\begin{tabular}{ll|cccc}
			\toprule
			&\multirow{2}{*}{Setting}& \multicolumn{2}{c}{DRSformer} & \multicolumn{2}{c}{Restormer} \\
			&& PSNR          & SSIM          & PSNR           & SSIM           \\
			\midrule
			\midrule
			(1)&Baseline & 32.42         & 0.9401        & 32.28               & 0.9388               \\
			(2)&(1)+PR      & 32.55         & 0.9410        &   32.45             &   0.9399             \\
			(3)&(2)+LL      & 32.72              & 0.9416        &   32.55             &  0.9403              \\
			(4)&(3)+$\mathcal{L}_{cycle}$    & 32.80         & 0.9420        &      32.60          &      0.9410        \\
			\bottomrule
		\end{tabular}
	\label{tab:table1}
\end{table}

\begin{figure}[t]
	\centering
	\includegraphics[width=0.99\linewidth]{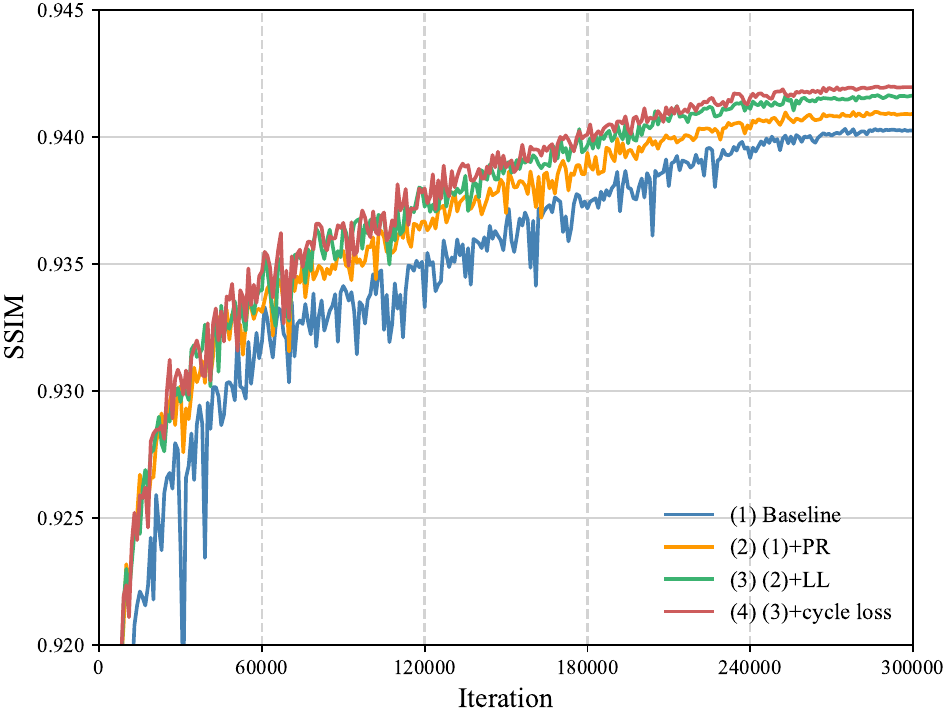}
	\caption{SSIM vs. number of iterations during the training phase. We record the SSIM values at certain iteration interval on Test-a.}
	\label{fig:fig5}
\end{figure}

\begin{table}[ht]
	\footnotesize
	\centering
	\caption{Ablation study of connecting the location-aware learning branch to different positions. Ei denotes level i of the encoder. The quantitative results are measured upon Test-a.}
		\begin{tabular}{c|cccc}
			\toprule
			& Baseline     & +LL   & +LL-E1 & +LL-E1234\\
			\midrule
			\midrule
			PSNR      & 32.42      & 32.56  & 32.50 & 32.22 \\
			SSIM      & 0.9401    & 0.9408 &  0.9403 & 0.9378 \\
			\bottomrule 
		\end{tabular}%
	\label{tab:table2}
\end{table}

\begin{table}[ht]
	\footnotesize
	\centering
	\caption{Ablation study of different supervisions on the learned location map $\textbf{M}$. The \# Param. (Tr./Te.) is obtained by statistical analysis of the network structure during the training/testing phase.}
		\begin{tabular}{c|ccc}
			\toprule
			& Baseline     & +LL(Diff)   & +LL(Binary)\\
			\midrule
			\midrule
			PSNR      & 32.42      & 32.56  & 32.13 \\
			SSIM      & 0.9401    & 0.9408 &  0.9401 \\
			\# Param. (Tr.) & 2,069,778       & 2,145,234  & 2,144,082   \\
			\# Param. (Te.) & 2,069,778       & 2,069,778  & 2,069,778   \\
			\bottomrule 
		\end{tabular}%
	\label{tab:table3}
\end{table}

\begin{figure}[h]
	\centering
	\includegraphics[width=0.99\linewidth]{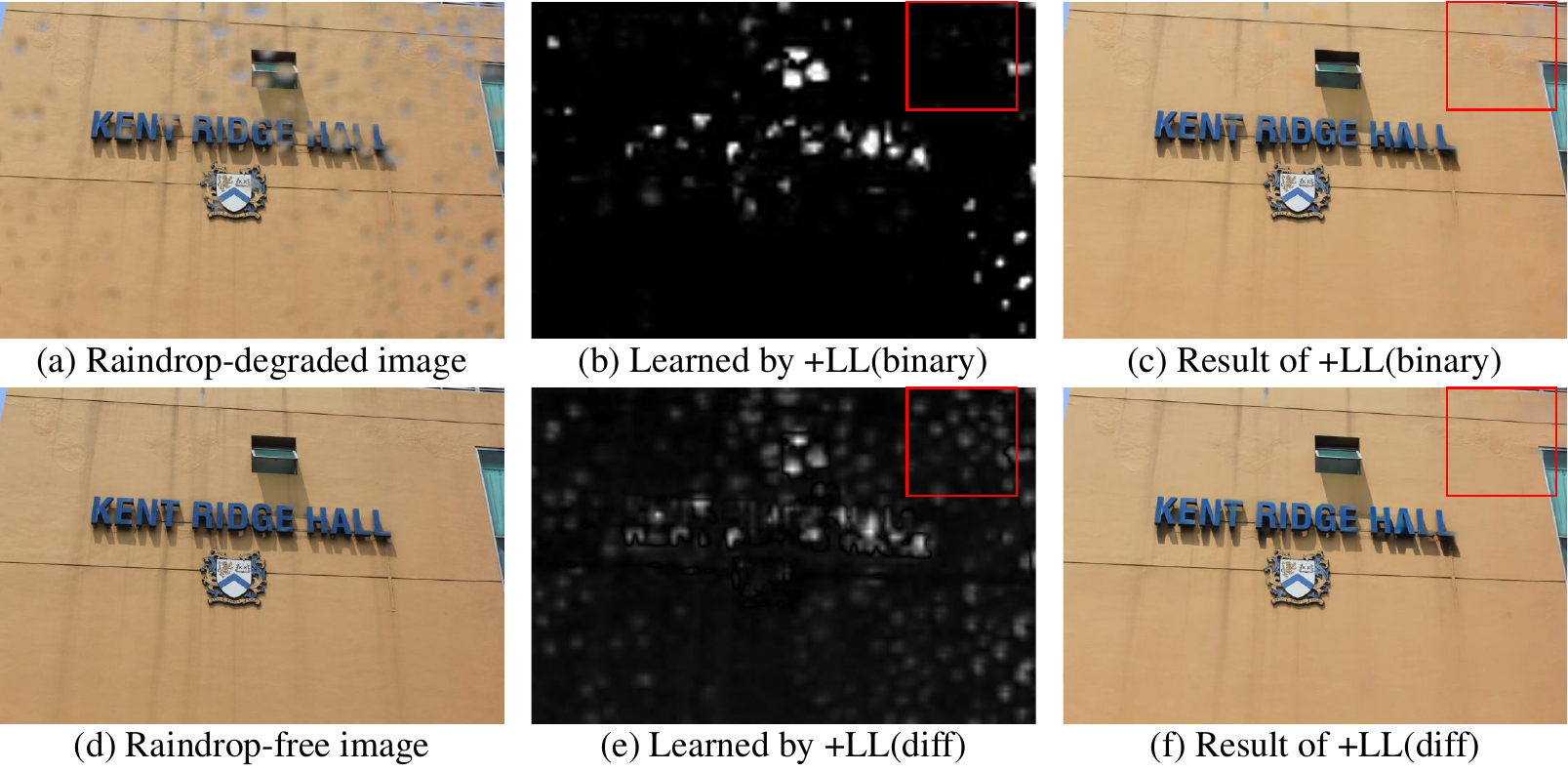}
	\caption{Visualization of learned masks and results of +LL(diff) and +LL(binary). Since the learned mask of +LL(diff) contains negative values, we adopt absolute values for better visualization.}
	\label{fig:fig6}
\end{figure}

For convenience, we train these models by setting the number of blocks deployed in each level $L_{1:4}$ to $\{1,2,2,4\}$.
The number of heads in self-attention at different levels are set to the values of $\{1,2,2,4\}$.
These settings keep consistent with the entire ablation study.
Although metrics are lower than the completely trained models reported in Table~\ref{tab:benchmark}, the trends and values are consistent and meaningful.

Table~\ref{tab:table1} shows the quantitative results of these four models upon Test-a.
Taking DRSformer \cite{Chen2023CVPR-DRSformer} as an example, physics-based reconstruction scheme, location-aware learning branch, and cycle loss function $\mathcal{L}_{cycle}$ bring 0.13 dB, 0.17 dB, and 0.08 dB improvements in terms of PSNR value respectively.
Similar results can be observed by taking Restormer \cite{Zamir2022CVPR-Restormer} as an example.
Unless otherwise specified, the Transformer block from DRSformer \cite{Chen2023CVPR-DRSformer} will be utilized throughout subsequent ablation study.
We meticulously document the rising trend of the evaluation metrics on the testing dataset for these four models during the training process.
Figure~\ref{fig:fig5} records the SSIM values of each iteration on Test-a.
It visually demonstrates the effectiveness of the designs within our LLPR framework.

\subsubsection{Position of location-aware learning branch}
In our implementation, the location-aware learning branch is deployed in the level 4 of the encoder (i.e., the second column of Table~\ref{tab:table2}).
We perform ablation study by deploying it in the level 1 of the encoder (the third column of Table~\ref{tab:table2}) and in all levels of the encoder (the final column of Table~\ref{tab:table2}).

We observe that the best performance is achieved when deploying in level 4, as it contains richest features.
Deploying across all four levels results in a performance drop, which we argue is attributed to the potential disruption in gradient back-propagation caused by the presence of multiple $\mathcal{L}_{loc}$ during the training process.

\subsubsection{Supervision of the learned location map $\textbf{M}$}
In our implementation, the difference between $\textbf{GT}$ and $\textbf{I}$ is utilized as the label to supervise the learned location map $\textbf{M}$ (it is denoted as +LL(diff)).
Another choice is to employ a binary mask \cite{Qian2018CVPR-AGAN} as the label (+LL(binary)).
We perform the ablation study of these two kinds of supervisions on our baseline model and the quantitative results on Test-a are summarized in Table~\ref{tab:table3}.

It is observed that the binary mask has a negative impact on the performance.
The binary mask only reflects the location characteristics and do not contain deep insightful information.
Furthermore, we visualize the learned location masks and processing results in Fig.~\ref{fig:fig6}.
We can observe that many slight raindrops are not recognized in the learned mask of +LL(binary), which subsequently leads to unsatisfactory raindrop removal performance (highlighted in the red rectangular region).

\begin{figure}[b]
	\centering
	\includegraphics[width=0.99\linewidth]{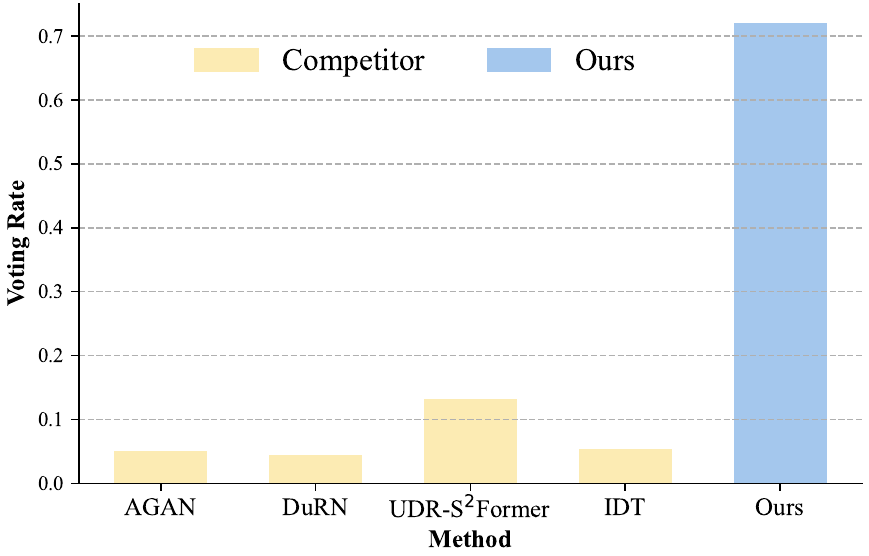}
	\caption{The user study result on our Test-wild dataset. We compare our LLPR(DRSformer) with 4 competitors.}
	\label{fig:fig7}
\end{figure}

\subsubsection{Location learning styles}
In our implementation, the location-aware learning branch is connected to level 4 of the encoder, which is independent from the restoration network structure.
We further perform ablation study with AGAN style \cite{Qian2018CVPR-AGAN} (i.e., setting 2 in Fig.~\ref{fig:fig1}) by inserting a sub-network (containing Res-LSTM blocks \cite{He2016CVPR-Resnet,Shi2015NIPS-LSTM}) before the restoration network structure.
The quantitative results on Test-a are summarized in Table~\ref{tab:table5-sm}.

\begin{table}[ht]
	\footnotesize
	\centering
	\caption{Ablation study of different location learning styles. The \# Param. (Tr./Te.) is obtained by statistical analysis of the network structure during the training/testing phase.}
		\begin{tabular}{c|ccc}
			\toprule
			& Baseline     & +LL   & +AGAN style\\
			\midrule
			\midrule
			PSNR      & 32.42      & 32.56  & 32.50 \\
			SSIM      & 0.9401    & 0.9408 &  0.9401 \\
			\# Param. (Tr.) & 2,069,778       & 2,145,234  & 2,237,731   \\
			\# Param. (Te.) & 2,069,778       & 2,069,778  & 2,237,731   \\
			Memory (GB)&  $\sim$8       &$\sim$8   &$\sim$12    \\
			\bottomrule 
		\end{tabular}%
	\label{tab:table5-sm}
\end{table}

It is observed that both styles can boost the performance in terms of PSNR.
However, our location-aware learning branch introduces only a small number of parameters during the training phase, which will not affect the testing phase at all.
Meanwhile, it also occupies significantly less GPU memory (during training) than the AGAN style.

\begin{table*}[ht]
	\footnotesize
	\centering
	\caption{Benchmark results of various raindrop removal methods on Test-a and Test-b datasets. We report PSNR, SSIM, number of parameters (\# Param.), and number of floating-point operations (\# FLOPs) to perform comprehensive comparisons. Note that \# FLOPs is measured on color images with $256\times 256$ resolution. \textbf{Bold} represents the state-of-the-art performance.}
	\begin{tabular}{c|c|c|cc|cc|cc}
		\toprule
		\multirow{2}{*}{Type} & \multirow{2}{*}{Method} & \multirow{2}{*}{Venue} & \multicolumn{2}{c|}{Test-a} & \multicolumn{2}{c|}{Test-b} & \multicolumn{2}{c}{Overhead}\\
		& & & PSNR & SSIM & PSNR & SSIM & \# Param. (M) & \# FLOPs (G) \\
		\midrule
		\midrule
		\multirow{3}{*}{Restoration} & DuRN \cite{Liu2019CVPR-DuRN} & CVPR'19 & 31.24 & 0.9259 & 25.32 & 0.8173 & 10.18 & 46.46 \\
		& MAXIM \cite{Tu2022CVPR-MAXIM} & CVPR'22 & 31.87 & 0.9352 & 25.74 & 0.8270 & 14.10 & 216.00 \\
		& Histoformer \cite{Sun2024ECCV-Histogram} &ECCV'24  & 33.06 & 0.9441 & - & - & - & - \\
		\midrule
		\multirow{4}{*}{Multi-task} & All-in-One \cite{Li2020CVPR} & CVPR'20 & 31.12 & 0.9268 & - & - & 44.00 & - \\
		& TransWeather \cite{Valanarasu2022CVPR} & CVPR'22 & 30.17 & 0.9157 & - & - & 31.00 & - \\
		& WGWS-Net \cite{Zhu2023CVPR} & CVPR'23 & 32.38 & 0.9378 & - & - & 5.97 & 71.02 \\
		& WeatherDiff$_{64}$ \cite{Ozdenizci2023PAMI} & TPAMI'23 & 30.71 & 0.9312 & 26.13 & 0.8325 & 82.96 & - \\
		\midrule
		\multirow{7}{*}{Task-specific} & pix2pix \cite{pix2pix} & CVPR'17 & 26.79 & 0.8644 & 23.50 & 0.7150 & - & - \\
		& AGAN \cite{Qian2018CVPR-AGAN} & CVPR'18 & 31.57 & 0.9023 & 24.92 & 0.8090 & 6.24 & 76.50 \\
		& Quan's \cite{Quan2019ICCV} & ICCV'19 & 31.44 & 0.9263 & - & - & - & - \\
		& UMAN \cite{Shao2021TIP} & TIP'21 & 31.47 & 0.9235 & 25.35 & 0.8197 & - & - \\
		& CCN \cite{Quan2021CVPR-CCN} & CVPR'21 & 31.34 & 0.9290 & - & - & 3.75 & 245.85 \\
		& IDT \cite{Xiao2023PAMI} & TPAMI'23 & 31.87 & 0.9313 & 25.63 & 0.8245 & 16.00 & 61.90 \\
		& UDR-S$^{2}$Former \cite{Chen2023ICCV-UDR} & ICCV'23 & 32.64 & 0.9430 & 26.92 & 0.8317 & 8.53 & 21.58 \\
		\midrule
		& DRSformer \cite{Chen2023CVPR-DRSformer} & - & 32.61 & 0.9409 & 26.84 & 0.8319 & \textbf{3.72} & \textbf{16.33}\\
		& LLPR(DRSformer) & - & \textbf{33.12} & \textbf{0.9444} & \textbf{27.11} & \textbf{0.8371} & 3.72{\tiny +864} & 16.39 \\
		\midrule
		& Restormer \cite{Zamir2022CVPR-Restormer} & - & 32.68 & 0.9423 &26.74 & 0.8320 & \textbf{2.69} & \textbf{11.89} \\
		& LLPR(Restormer) & - & \textbf{33.11}  & \textbf{0.9445} & \textbf{27.03}  &\textbf{0.8365}  &2.69{\tiny +864}  &11.94  \\
		\bottomrule
	\end{tabular}
	\label{tab:benchmark}
\end{table*}

\begin{figure*}[h]
	\centering
	\includegraphics[width=0.99\linewidth]{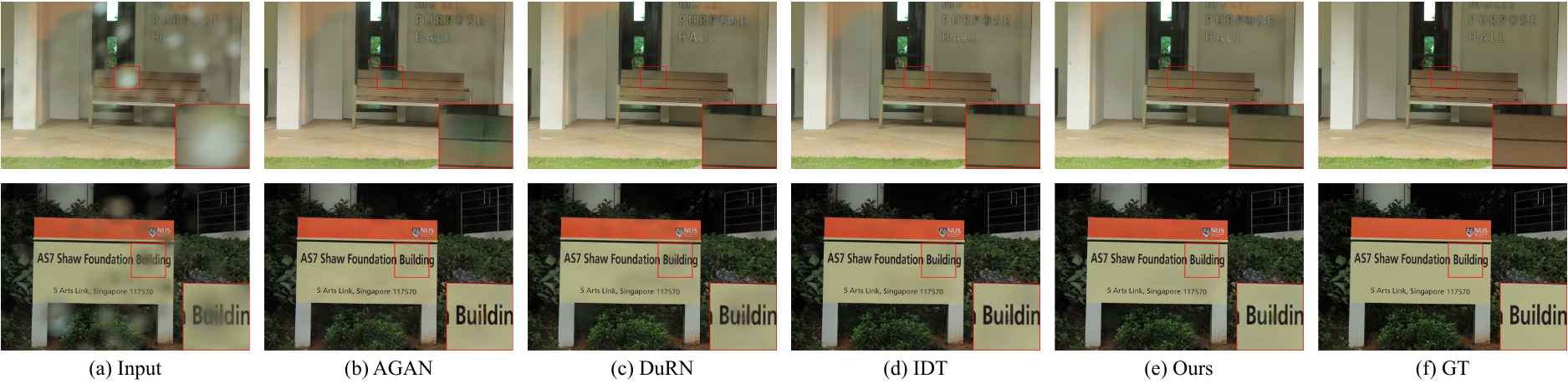}
	\caption{Visual comparisons of various methods on Test-a dataset. Please zoom in on screen for a better view.}
	\label{fig:fig8}
\end{figure*}

\begin{figure*}[!h]
	\centering
	\includegraphics[width=0.99\linewidth]{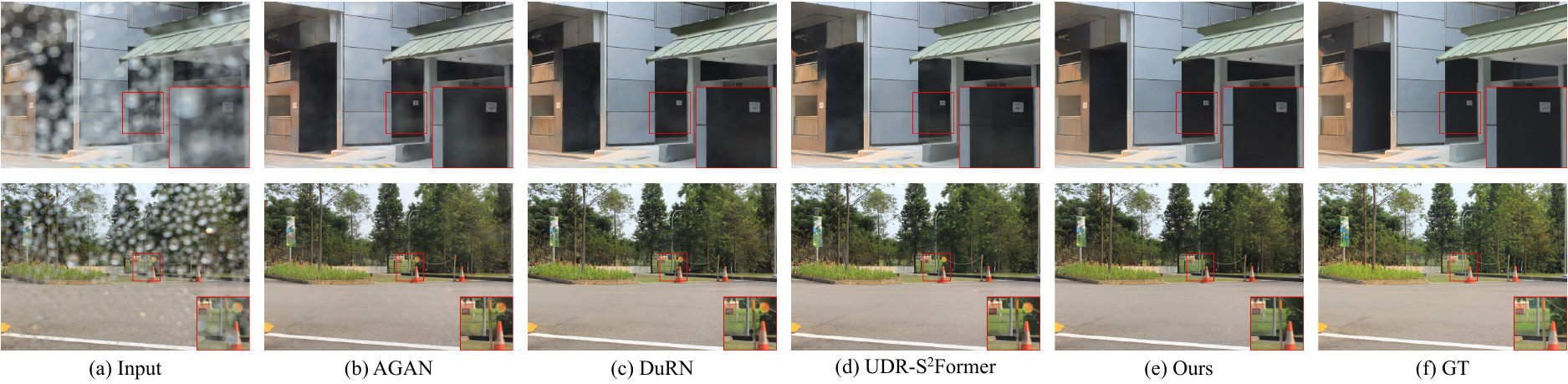}
	\caption{Visual comparisons of various methods on Test-b dataset. Please zoom in on screen for a better view.}
	\label{fig:fig9}
\end{figure*}

\subsection{Comparisons with state-of-the-art methods}
\subsubsection{Quantitative comparisons}
In this section, we train two variants based on our LLPR framework (select the basic Transformer block from DRSformer \cite{Chen2023CVPR-DRSformer} and Restormer \cite{Zamir2022CVPR-Restormer}) and compare them with seven task-specific methods including pip2pix \cite{pix2pix}, AGAN \cite{Qian2018CVPR-AGAN}, Quan's \cite{Quan2019ICCV}, UMAN \cite{Shao2021TIP}, CCN \cite{Quan2021CVPR-CCN}, IDT \cite{Xiao2023PAMI}, UDR-S$^2$Former \cite{Chen2023ICCV-UDR}, three image restoration methods including DuRN \cite{Liu2019CVPR-DuRN}, MAXIM \cite{Tu2022CVPR-MAXIM}, Histoformer \cite{Sun2024ECCV-Histogram} and four multi-task methods including All-in-One \cite{Li2020CVPR}, TransWeather \cite{Valanarasu2022CVPR}, WGWS-Net \cite{Zhu2023CVPR}, WeatherDiff \cite{Ozdenizci2023PAMI} on Test-a and Test-b datasets.
Table~\ref{tab:benchmark} shows quantitative evaluation results.
As we can see, both of LLPR(DRSformer) and LLPR(Restormer) achieve state-of-the-art performance.

In addition, we adopt number of parameters (\# Param.) and number of floating-point operations (\# FLOPs) as the major indicators of computational efficiency.
Compared with recent SOTA competitors, our methods can maintain the lowest \# Param. (please refer to Fig.~\ref{fig:fig2}) and \# FLOPs while achieving the best performance.
Note that \# FLOPs is measured on color images with $256\times 256$ resolution.

We conduct a user study to evaluate the proposed method (LLPR(DRSformer) is selected here) subjectively against other methods.
50 images from Test-wild are randomly selected for comparison.
We invite 10 experts with image restoration background as volunteers.
With the experiment, we refer to \cite{Wu2023CVPR-RIDCP} and follow their code of conduct.
Input image and the results generated by \{AGAN \cite{Qian2018CVPR-AGAN}, DuRN \cite{Liu2019CVPR-DuRN}, UDR-S$^2$Former \cite{Chen2023ICCV-UDR}, IDT \cite{Xiao2023PAMI}, and LLPR(DRSformer)\} are displayed to the observers group by group.
The observers are demanded to choose the best-performing after at least 10 seconds of observation.
Afterward, we statistic the percentage of certain method to be selected as the best-performing.
The voting rate in Fig.~\ref{fig:fig7} indicates that our LLPR(DRSformer) is more favored by these experts than competitors.

Since there is no aligned clean images for our collected Test-wild dataset, we also employ some no-reference image quality assessment (NRIQA) methods (e.g., NIQE \cite{NIQE}, MUSIQ \cite{MUSIQ}, PaQ-2-PiQ \cite{PaQ-2-PiQ}, HyperIQA \cite{HyberIQA}) to quantitatively evaluate the quality of raindrop removal results.
\{AGAN \cite{Qian2018CVPR-AGAN}, DuRN \cite{Liu2019CVPR-DuRN}, IDT \cite{Xiao2023PAMI}, and LLPR(DRSformer)\} are selected as the competitors, and the quantitative results are listed in Table~\ref{tab:table6-sm}.
Our method outperforms all compared methods across these evaluation metrics, which means the raindrop removal results are more aligned with the human visual perception.

\begin{table}[t]
	\footnotesize
	\centering
	\caption{Quantitative results of competing methods on Test-wild in terms of no-reference metrics. $\uparrow$ denotes that the higher value indicates the better performance. $\downarrow$ denotes that the lower value indicates the better performance. \textbf{Bold} represents the state-of-the-art performance.}
		\begin{tabular}{c|cccc}
			\toprule
			& NIQE$\downarrow$     & MUSIQ$\uparrow$  & PaQ-2-PiQ$\uparrow$  & HyperIQA$\uparrow$  \\
			\midrule
			\midrule
			Input                              &  4.758 & 59.91 & 66.66 & 0.5659   \\
			AGAN \cite{Qian2018CVPR-AGAN}      &  5.116 & 63.28 & 70.13 & 0.6226  \\
			DuRN \cite{Liu2019CVPR-DuRN}       &  4.630 & 63.80 & 70.67 & 0.6186  \\
			IDT \cite{Xiao2023PAMI}            &  3.334 & 64.87 & 70.95 & 0.6344  \\
			Ours                               &  \textbf{3.283} & \textbf{65.24} & \textbf{71.09} & \textbf{0.6514}  \\
			\bottomrule 
		\end{tabular}%
	\label{tab:table6-sm}
\end{table}

\begin{figure*}[ht]
	\centering
	\includegraphics[width=0.99\linewidth]{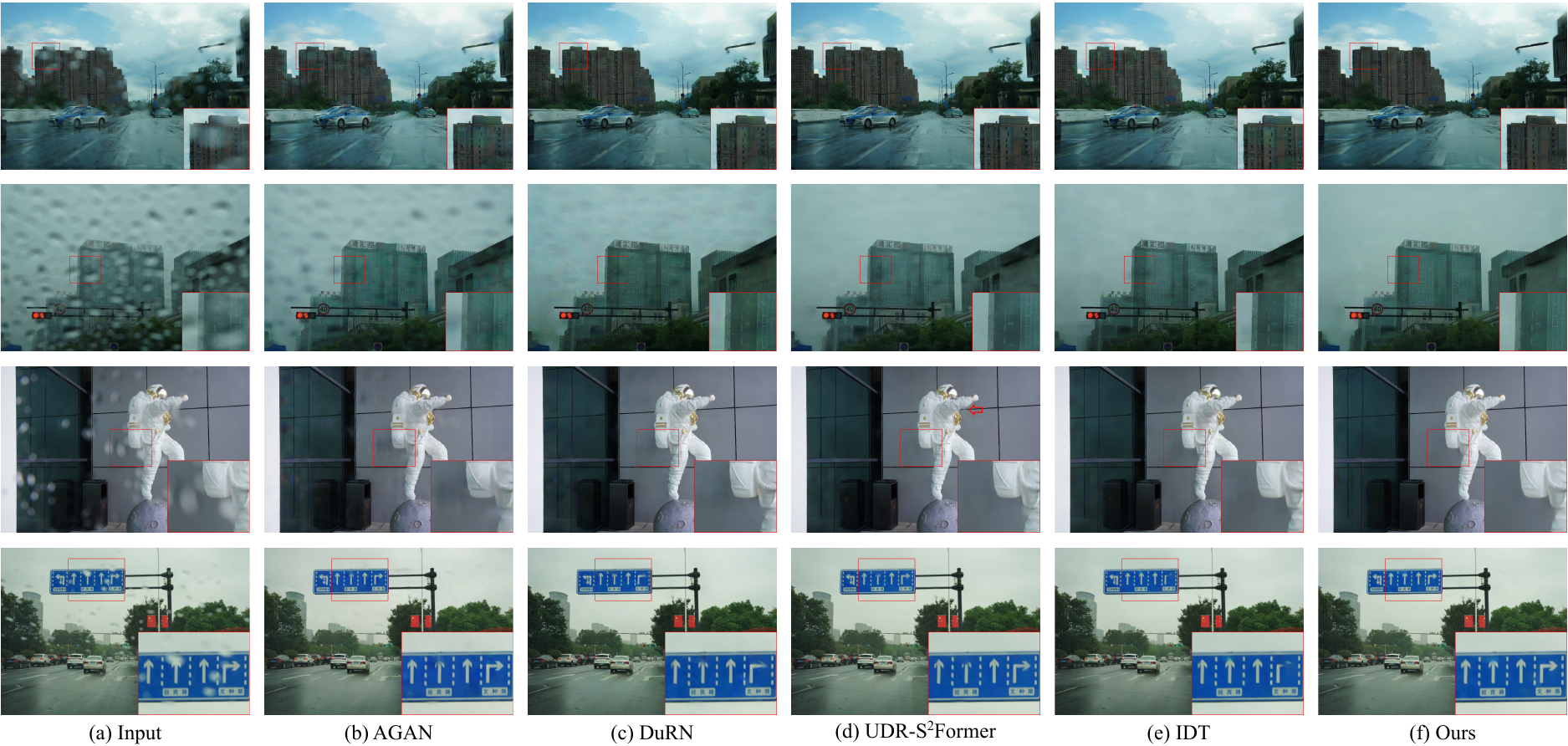}
	\caption{Visual comparisons of various methods on our collected Test-wild dataset. Please zoom in on screen for a better view.}
	\label{fig:fig10}
\end{figure*}

\subsubsection{Qualitative comparisons}
We perform visual comparisons between our LLPR(DRSformer) and previous SOTA methods on Test-a and Test-b.
As shown in Fig.~\ref{fig:fig8}, our method can recover sharper and clearer contours of the background scene. 
Among them, the visual results of IDT \cite{Xiao2023PAMI} contain some imperceptible block artifacts.
Similarly, as shown in Fig.~\ref{fig:fig9}, the results of our method are closest to the ground truth than the other alternatives.
Moreover, we also test on captured Test-wild, and the visual results are illustrated in Fig.~\ref{fig:fig10}.
We can clearly observe the priority of our LLPR in real-world rainy scenarios (astronaut and road sign).

\section{Conclusion}
\label{sec: conclusion}
In this paper, we propose our LLPR framework for SIRR problem, which is challenging and severely ill-posed.
Contrary to traditional approaches that primarily focus on modifying the architectures, several versatile techniques is proposed (location-aware learning branch, physics-based reconstruction scheme, cycle loss function) which can be equipped with existing structures to promote the performance.
We train two variants based on our LLPR framework by selecting the basic Transformer block in DRSformer \cite{Chen2023CVPR-DRSformer} and Restormer \cite{Zamir2022CVPR-Restormer}.
Extensive experiments demonstrate the effectiveness and generality of our LLPR framework, achieving superior performance against state-of-the-arts.



\bibliographystyle{IEEEtran}
\bibliography{reference}

\begin{thebibliography}{10}
\providecommand{\url}[1]{#1}
\csname url@samestyle\endcsname
\providecommand{\newblock}{\relax}
\providecommand{\bibinfo}[2]{#2}
\providecommand{\BIBentrySTDinterwordspacing}{\spaceskip=0pt\relax}
\providecommand{\BIBentryALTinterwordstretchfactor}{4}
\providecommand{\BIBentryALTinterwordspacing}{\spaceskip=\fontdimen2\font plus
\BIBentryALTinterwordstretchfactor\fontdimen3\font minus
  \fontdimen4\font\relax}
\providecommand{\BIBforeignlanguage}[2]{{%
\expandafter\ifx\csname l@#1\endcsname\relax
\typeout{** WARNING: IEEEtran.bst: No hyphenation pattern has been}%
\typeout{** loaded for the language `#1'. Using the pattern for}%
\typeout{** the default language instead.}%
\else
\language=\csname l@#1\endcsname
\fi
#2}}
\providecommand{\BIBdecl}{\relax}
\BIBdecl

\bibitem{Eigen2013}
D.~Eigen, D.~Krishnan, and R.~Fergus, ``{Restoring an image taken through a
  window covered with dirt or rain},'' in \emph{ICCV}, 2013, pp. 633--640.

\bibitem{Quan2019ICCV}
Y.~Quan, S.~Deng, Y.~Chen, and H.~Ji, ``{Deep learning for seeing through
  window with raindrops},'' in \emph{ICCV}, 2019, pp. 2463--2471.

\bibitem{Shao2021Neurocom}
M.~Shao, L.~Li, H.~Wang, and D.~Meng, ``{Selective generative adversarial
  network for raindrop removal from a single image},'' \emph{Neurocomputing},
  vol. 426, pp. 265--273, 2021.

\bibitem{Guo2020TIP}
Y.~Guo, J.~Chen, X.~Ren, A.~Wang, and W.~Wang, ``{Joint Raindrop and Haze
  Removal from a Single Image},'' \emph{IEEE Transactions on Image Processing},
  vol.~29, pp. 9508--9519, 2020.

\bibitem{Luo2021TCSVT}
W.~Luo, J.~Lai, and X.~Xie, ``{Weakly Supervised Learning for Raindrop Removal
  on a Single Image},'' \emph{IEEE Transactions on Circuits and Systems for
  Video Technology}, vol.~31, no.~5, pp. 1673--1683, 2021.

\bibitem{You2016}
S.~You, R.~T. Tan, R.~Kawakami, Y.~Mukaigawa, and K.~Ikeuchi, ``{Adherent
  Raindrop Modeling, Detection and Removal in Video},'' \emph{IEEE Transactions
  on Pattern Analysis {\&} Machine Intelligence}, vol.~38, no.~9, pp.
  1721--1733, 2016.

\bibitem{Qian2018CVPR-AGAN}
R.~Qian, R.~T. Tan, W.~Yang, J.~Su, and J.~Liu, ``{Attentive Generative
  Adversarial Network for Raindrop Removal from a Single Image},'' in
  \emph{CVPR}, 2018, pp. 2482--2491.

\bibitem{Peng2020PRL}
J.~Peng, Y.~Xu, T.~Chen, and Y.~Huang, ``{Single-image raindrop removal using
  concurrent channel-spatial attention and long-short skip connections},''
  \emph{Pattern Recognition Letters}, vol. 131, pp. 121--127, 2020.

\bibitem{Shao2021TIP}
M.~W. Shao, L.~Li, D.~Y. Meng, and W.~M. Zuo, ``{Uncertainty Guided Multi-Scale
  Attention Network for Raindrop Removal from a Single Image},'' \emph{IEEE
  Transactions on Image Processing}, vol.~30, pp. 4828--4839, 2021.

\bibitem{Chen2023ICCV-UDR}
S.~Chen, T.~Ye, J.~Bai, E.~Chen, J.~Shi, and L.~Zhu, ``{Sparse Sampling
  Transformer with Uncertainty-Driven Ranking for Unified Removal of Raindrops
  and Rain Streaks},'' in \emph{ICCV}, 2023, pp. 13\,106--13\,117.

\bibitem{Liu2019CVPR-DuRN}
X.~Liu, M.~Suganuma, Z.~Sun, and T.~Okatani, ``Dual residual networks
  leveraging the potential of paired operations for image restoration,'' in
  \emph{CVPR}, 2019, pp. 7007--7016.

\bibitem{Tu2022CVPR-MAXIM}
Z.~Tu, H.~Talebi, H.~Zhang, F.~Yang, P.~Milanfar, A.~C. Bovik, and Y.~Li,
  ``{MAXIM:} multi-axis {MLP} for image processing,'' in \emph{CVPR}, 2022, pp.
  5759--5770.

\bibitem{Sun2024ECCV-Histogram}
S.~Sun, W.~Ren, X.~Gao, R.~Wang, and X.~Cao, ``{Restoring Images in Adverse
  Weather Conditions via Histogram Transformer},'' in \emph{ECCV}, 2024.

\bibitem{He2016CVPR-Resnet}
K.~He, X.~Zhang, S.~Ren, and J.~Sun, ``{Deep Residual Learning for Image
  Recognition},'' in \emph{CVPR}, 2016, pp. 770--778.

\bibitem{Shi2015NIPS-LSTM}
X.~Shi, Z.~Chen, H.~Wang, D.-Y. Yeung, W.-K. Wong, and W.-c. Woo,
  ``Convolutional lstm network: A machine learning approach for precipitation
  nowcasting,'' \emph{NeurIPS}, vol.~28, 2015.

\bibitem{Chen2023CVPR-DRSformer}
X.~Chen, H.~Li, M.~Li, and J.~Pan, ``{Learning A Sparse Transformer Network for
  Effective Image Deraining},'' in \emph{CVPR}, 2023, pp. 5896--5905.

\bibitem{Zamir2022CVPR-Restormer}
S.~W. Zamir, A.~Arora, S.~Khan, M.~Hayat, F.~S. Khan, and M.-H. Yang,
  ``{Restormer: Efficient Transformer for High-Resolution Image Restoration},''
  in \emph{CVPR}, 2022, pp. 5728--5739.

\bibitem{Li2020CVPR}
R.~Li, R.~T. Tan, and L.~Cheong, ``All in one bad weather removal using
  architectural search,'' in \emph{CVPR}, 2020, pp. 3172--3182.

\bibitem{Valanarasu2022CVPR}
J.~M.~J. Valanarasu, R.~Yasarla, and V.~M. Patel, ``Transweather:
  Transformer-based restoration of images degraded by adverse weather
  conditions,'' in \emph{CVPR}, 2022, pp. 2343--2353.

\bibitem{Zhu2023CVPR}
Y.~Zhu, T.~Wang, X.~Fu, X.~Yang, X.~Guo, J.~Dai, Y.~Qiao, and X.~Hu, ``Learning
  weather-general and weather-specific features for image restoration under
  multiple adverse weather conditions,'' in \emph{CVPR}, 2023, pp.
  21\,747--21\,758.

\bibitem{Huang2024TMM-WaveDM}
Y.~Huang, J.~Huang, J.~Liu, M.~Yan, Y.~Dong, J.~Lv, C.~Chen, and S.~Chen,
  ``{WaveDM: Wavelet-Based Diffusion Models for Image Restoration},''
  \emph{IEEE Transactions on Multimedia}, vol.~26, pp. 7058--7073, 2024.

\bibitem{Quan2021CVPR-CCN}
R.~Quan, X.~Yu, Y.~Liang, and Y.~Yang, ``Removing raindrops and rain streaks in
  one go,'' in \emph{CVPR}, 2021, pp. 9147--9156.

\bibitem{Xiao2023PAMI}
J.~Xiao, X.~Fu, A.~Liu, F.~Wu, and Z.-J. Zha, ``Image de-raining transformer,''
  \emph{IEEE Transactions on Pattern Analysis and Machine Intelligence},
  vol.~45, no.~11, pp. 12\,978--12\,995, 2023.

\bibitem{Chen2019CVPRW}
D.~Chen, Z.~He, A.~Sun, J.~Yang, Y.~Cao, Y.~Cao, S.~Tang, and M.~Y. Yang,
  ``{Orientation-aware Deep Neural Network for Real Image Super-Resolution},''
  in \emph{CVPR Workshop}, 2019.

\bibitem{Shi2016CVPR}
W.~Shi, J.~Caballero, F.~Huszar, J.~Totz, A.~P. Aitken, R.~Bishop, D.~Rueckert,
  and Z.~Wang, ``{Real-Time Single Image and Video Super-Resolution Using an
  Efficient Sub-Pixel Convolutional Neural Network},'' in \emph{CVPR}, 2016,
  pp. 1874--1883.

\bibitem{Liu2024Neurocom}
Y.~Liu, Z.~Gao, T.~Mei, and H.~Yi, ``{Single image deraindrop leveraging
  luminance priors and context aggregation},'' \emph{Neurocomputing}, vol. 598,
  p. 127971, sep 2024.

\bibitem{Li2019CVPR}
R.~Li, L.-F. Cheong, and R.~T. Tan, ``Heavy rain image restoration: Integrating
  physics model and conditional adversarial learning,'' in \emph{CVPR}, 2019,
  pp. 1633--1642.

\bibitem{Zhang2018CVPR}
H.~Zhang and V.~M. Patel, ``{Densely Connected Pyramid Dehazing Network},'' in
  \emph{CVPR}, 2018, pp. 3194--3203.

\bibitem{Wang2004TIP-SSIM}
Z.~Wang, A.~C. Bovik, H.~R. Sheikh, and E.~P. Simoncelli, ``{Image quality
  assessment: from error visibility to structural similarity.}'' \emph{IEEE
  Transactions on Image Processing}, vol.~13, no.~4, pp. 600--612, apr 2004.

\bibitem{Paszke2019NIPS-PyTorch}
A.~Paszke, S.~Gross, F.~Massa, A.~Lerer, J.~Bradbury, G.~Chanan, T.~Killeen,
  Z.~Lin, N.~Gimelshein, L.~Antiga \emph{et~al.}, ``Pytorch: An imperative
  style, high-performance deep learning library,'' \emph{Advances in neural
  information processing systems}, vol.~32, 2019.

\bibitem{AdamW}
I.~Loshchilov, F.~Hutter \emph{et~al.}, ``Fixing weight decay regularization in
  adam,'' \emph{arXiv preprint arXiv:1711.05101}, 2017.

\bibitem{He2019CVPR-Bag}
T.~He, Z.~Zhang, H.~Zhang, Z.~Zhang, J.~Xie, and M.~Li, ``Bag of tricks for
  image classification with convolutional neural networks,'' in \emph{CVPR},
  2019, pp. 558--567.

\bibitem{Ozdenizci2023PAMI}
O.~{\"{O}}zdenizci and R.~Legenstein, ``{Restoring Vision in Adverse Weather
  Conditions With Patch-Based Denoising Diffusion Models},'' \emph{IEEE
  Transactions on Pattern Analysis and Machine Intelligence}, vol.~45, no.~8,
  pp. 10\,346--10\,357, aug 2023.

\bibitem{pix2pix}
P.~Isola, J.-Y. Zhu, T.~Zhou, and A.~A. Efros, ``Image-to-image translation
  with conditional adversarial networks,'' in \emph{CVPR}, 2017, pp.
  1125--1134.

\bibitem{Wu2023CVPR-RIDCP}
R.-Q. Wu, Z.-P. Duan, C.-L. Guo, Z.~Chai, and C.~Li, ``{RIDCP: Revitalizing
  Real Image Dehazing via High-Quality Codebook Priors},'' in \emph{CVPR},
  2023, pp. 22\,282--22\,291.

\bibitem{NIQE}
A.~Mittal, R.~Soundararajan, and A.~C. Bovik, ``{Making a “Completely
  Blind” Image Quality Analyzer},'' \emph{IEEE Signal Processing Letters},
  vol.~20, no.~3, pp. 209--212, mar 2013.

\bibitem{MUSIQ}
J.~Ke, Q.~Wang, Y.~Wang, P.~Milanfar, and F.~Yang, ``Musiq: Multi-scale image
  quality transformer,'' in \emph{ICCV}, 2021, pp. 5148--5157.

\bibitem{PaQ-2-PiQ}
Z.~Ying, H.~Niu, P.~Gupta, D.~Mahajan, D.~Ghadiyaram, and A.~Bovik, ``From
  patches to pictures (paq-2-piq): Mapping the perceptual space of picture
  quality,'' in \emph{CVPR}, 2020, pp. 3575--3585.

\bibitem{HyberIQA}
S.~Su, Q.~Yan, Y.~Zhu, C.~Zhang, X.~Ge, J.~Sun, and Y.~Zhang, ``{Blindly Assess
  Image Quality in the Wild Guided by a Self-Adaptive Hyper Network},'' in
  \emph{CVPR}, 2020, pp. 3664--3673.

\end{thebibliography}

\end{document}